\documentclass[a4paper,fleqn]{cas-dc}

\usepackage[numbers]{natbib}
\usepackage{amsmath,amssymb,amsfonts}
\usepackage{multirow}
\usepackage{graphicx}
\usepackage{rotating}
\usepackage{multicol}
\usepackage{amsmath}
\usepackage{float}
\usepackage{array}
\usepackage{caption}
\usepackage{subcaption}
\usepackage{siunitx}
\usepackage{float}
\usepackage{hyperref}
\usepackage{xcolor}
\usepackage{multirow}
\usepackage{graphicx}
\usepackage{color, soul, colortbl}
\usepackage{nicematrix}
\usepackage{colortbl} 

\DeclareUnicodeCharacter{2061}{}

\begin{document}

\title{\LARGE MuLA-GAN: Multi-Level Attention GAN for Enhanced Underwater Visibility}


\author[1]{Ahsan B. Bakht}[orcid=0000-0002-9079-0960]
\author[2]{Zikai Jia}[orcid=0000-0001-9789-8483]
\author[1]{Muhayy ud Din}[orcid=0000-0001-6214-1077]
\author[1]{Waseem Akram}[orcid=0000-0002-7401-5120]
\author[1]{Lyes Saad Soud}[orcid=0000-0003-4445-3135]
\author[1]{Lakmal Seneviratne}[orcid=0000-0001-6405-8402]
\author[2]{Defu Lin}
\author[2]{Shaoming He}[orcid=0000-0001-6432-5187]
\author[1]{Irfan Hussain}[orcid=0000-0003-2759-0306]

\affiliation[1]{Khalifa University Center for Autonomous Robotic Systems (KUCARS), Khalifa University, United Arab Emirates}
\affiliation[2]{ School of Aerospace Engineering, Beijing Institute of Technology, China.}
\tnotetext[1]{First two authors has equal contribution}

\shorttitle {Underwater Image Enhancement}

\begin{abstract}
The underwater environment presents unique challenges, including color distortions, reduced contrast, and blurriness, hindering accurate analysis. In this work, we introduce MuLA-GAN, a novel approach that leverages the synergistic power of Generative Adversarial Networks (GANs) and Multi-Level Attention mechanisms for comprehensive underwater image enhancement.
The integration of Multi-Level Attention within the GAN architecture significantly enhances the model's capacity to learn discriminative features crucial for precise image restoration. By selectively focusing on relevant spatial and multi-level features, our model excels in capturing and preserving intricate details in underwater imagery, essential for various applications.
Extensive qualitative and quantitative analyses on diverse datasets, including UIEB test dataset, UIEB challenge dataset, U45, and UCCS dataset, highlight the superior performance of MuLA-GAN compared to existing state-of-the-art methods. Experimental evaluations on a specialized dataset tailored for bio-fouling and aquaculture applications demonstrate the model's robustness in challenging environmental conditions.
On the UIEB test dataset, MuLA-GAN achieves exceptional PSNR $(25.59)$ and SSIM $(0.893)$ scores, surpassing Water-Net, the second-best model, with scores of $24.36$ and $0.885$, respectively. This work not only addresses a significant research gap in underwater image enhancement but also underscores the pivotal role of Multi-Level Attention in enhancing GANs, providing a novel and comprehensive framework for restoring underwater image quality.
\end{abstract}


\maketitle

\section{Introduction}
In recent years, the vast potential of our oceans, from abundant resources to environmental significance, has garnered increasing attention. The exploration and exploitation of marine resources have become focal points for researchers. However, the acquisition of clear and detailed underwater images faces significant challenges, including blurriness, color distortion, and degradation caused by light interference and complex underwater backgrounds, or due to extreme weather conditions as observed by \cite{Zhang1},\cite{ahmed2023vision}. These challenges lead to the loss of crucial details, making visual tasks such as feature extraction and target detection exceptionally demanding.

Addressing these challenges, the field of underwater image enhancement strives to elevate the quality of underwater images, unlocking more effective utilization of the visual information available \cite{guo2019underwater}. This enhancement holds significant practical applications, spanning target detection, marine environment monitoring, wreck detection, and military operations. Applying image processing techniques to improve these images can greatly benefit various high-level computer vision tasks.

The quality of underwater images profoundly influences the efficacy of underwater applications. Factors such as limited visibility, uneven lighting, noise, and color distortion contribute to the difficulties in obtaining high-quality underwater images \cite{azmi2019natural}. Underwater image processing broadly falls into two categories: image restoration and image enhancement \cite{raveendran2021underwater}. While image restoration aims to recover the original content of underwater images by addressing issues like blurring and color distortion, image enhancement techniques focus on improving visual appeal without necessarily adhering to a physical model or aiming for an accurate restoration.
\begin{figure*}[t]
    \centering
    \includegraphics[width=12cm]{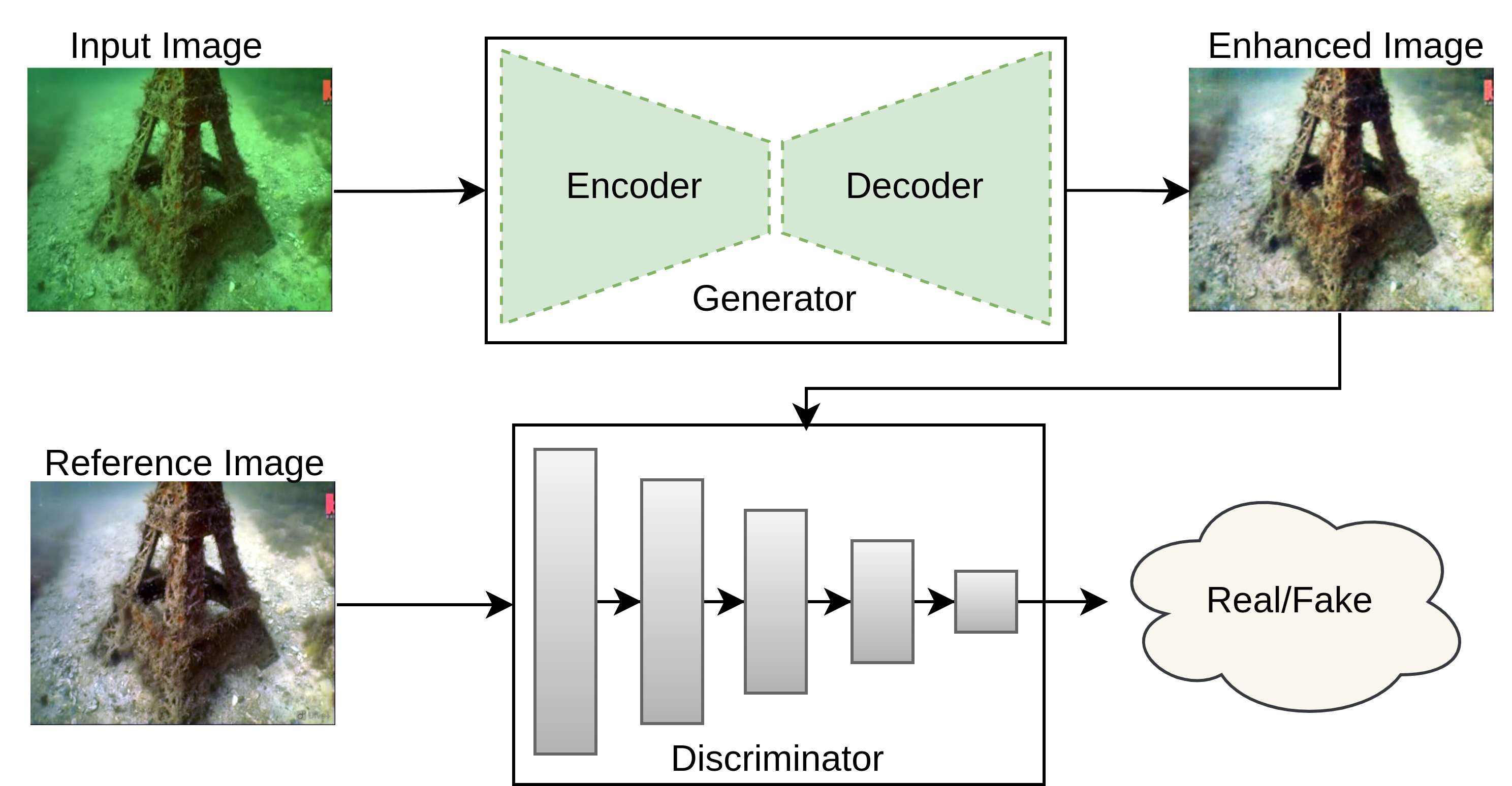}
    \caption{Proposed Dehazing Model Training Process: An input image transforms within our generator network, integral to a GAN model. The resultant output image undergoes rigorous evaluation by a discriminator network, in conjunction with a reference image, for authenticity assessment. The generator refines its output based on this evaluation, iteratively crafting more realistic images. Its objective is to strategically deceive the discriminator while aligning with the characteristics of the reference image, thereby mastering the nuances of image enhancement.}
    \label{fig:Main_Diagram}
\end{figure*}
Traditional underwater image enhancement methods often prioritize visually pleasing results based on subjective criteria. These methods, simpler and faster than model-based approaches, can produce appealing results but may alter image features, impacting tasks such as object detection and segmentation \cite{anwar2020diving}. In contrast, model-based approaches leveraging deep learning architectures have demonstrated superiority, particularly in preserving features accurately. However, existing GAN-based approaches, including UGAN \cite{d2}, WaterGAN \cite{d1}, and FunieGAN \cite{d5}, while promising in image enhancement and restoration, introduce noise in the generated images. Conventional convolutional networks like UWCNN \cite{d6} and WaterNet \cite{d7} exhibit color correction deviations, limiting their capabilities across diverse underwater scenarios.

In response to the strengths and limitations of existing models, this paper introduces MuLA-GAN, a novel GAN-based approach that integrates a spatio-channel attention mechanism into the generator architecture. This integration significantly enhances the model's ability to perform effective image enhancement and restoration. The primary contributions of this paper are:
\begin{itemize}
    \item Introducing MuLA-GAN, a novel and effective image enhancement method based on a GANs model that incorporates a spatio-channel attention mechanism into the generator architecture, significantly improving the model's effectiveness.
    \item Conducting an extensive qualitative and quantitative analysis on publicly available datasets, including UIEB, U45, and UCCS. This analysis focuses on the visual aspects of image enhancement and includes a detailed comparison of the proposed method with state-of-the-art approaches across diverse scenes and scenarios within these datasets.
    \item Performing a comprehensive analysis on a distinct dataset collected in our laboratory, primarily focused on bio-fouling and aquaculture applications. This analysis provides insights into the model's performance in diverse environmental conditions.
\end{itemize}

\section{Related Work}\label{sec:r-work}
Image enhancement plays a crucial role in improving visual quality and extracting meaningful information from images. Over the years, researchers have explored various methodologies to enhance images, with two prominent streams of investigation emerging. One branch primarily focuses on leveraging traditional image processing techniques, while the other harnesses the power of deep learning algorithms. In this section, we review some of the currently available solutions making use of classical and deep learning-based approaches.

\subsection{Classical image enhancement techniques}

The classical approaches to image enhancement through image processing often involve manually designed techniques. For improving the overall visibility of low-light images, contrast enhancement methods such as histogram equalization \cite{hummel1975image}, contrast-limited adaptive histogram equalization (CLAHE) \cite{reza2004realization}, and gamma correction are commonly utilized. These techniques aim to enhance the global contrast and visibility of the images. The gray-world assumption and white balancing are also widely used traditional color correction methods that are commonly used to adjust the color and saturation of images. These fundamental image-processing techniques have been traditionally utilized for image enhancement and are now being increasingly adapted for underwater scenarios. By applying these techniques to underwater images, researchers aim to improve color accuracy and enhance the visual quality of underwater scenes.

Research studies have explored various image processing techniques for improving the quality of underwater images. The contrast stretching and fuzzy HE methods \cite{karam2013enhancement8} have been used by researchers to enhance underwater image quality. Additionally, the inversion of light absorption \cite{petit2009underwater9} has been proposed as a solution to the underwater color correction problem. Fusion enhancement strategies \cite{ancuti2012enhancing10} have also been designed, involving the weighted fusion of images after WB and HE. The dark channel prior (DCP) \cite{he2010single11}, originally developed for dehazing, has found extensive use in underwater image recovery. An underwater-specific variation of the DCP, known as the underwater dark channel prior (UDCP) \cite{drews2016underwater12}, utilizes the minima of green and blue channels to compute dark channel images. The generalized dark channel prior (GDCP) \cite{peng2018generalization13} incorporates adaptive color correction into image formation models for image recovery. Submerged dark channel prior (SDCP) \cite{chang2018single14} has been proposed for removing scattering and estimating the conduction equation in image restoration. Real-time underwater application benchmarks have been established by Perez et al. \cite{perez2017benchmarking15} for evaluating DCP-related models. The Jaffe-McGlamery model \cite{mcglamery1975computer16},\cite{mcglamery1980computer17} is another commonly used underwater imaging model. A physically accurate model based on the Jaffe-McGlamery model has been proposed by Akkaynak et al. \cite{akkaynak2017space18}, \cite{akkaynak2018revised19}, where the attenuation coefficient characterizes target distance and reflection. Utilizing RGBD images, the Sea-thru method \cite{akkaynak2019sea20} estimates backscattering and distance-dependent decay coefficients, which play vital roles in enhancing underwater image quality. These image-processing techniques aim to enhance contrast and improve the overall image quality of underwater scenes. However, individual techniques are often developed based on specific assumptions or for particular tasks, limiting their suitability for addressing general underwater enhancement scenarios with significant variability.
\begin{figure*}[t]
    \centering
    \includegraphics[width=\textwidth]{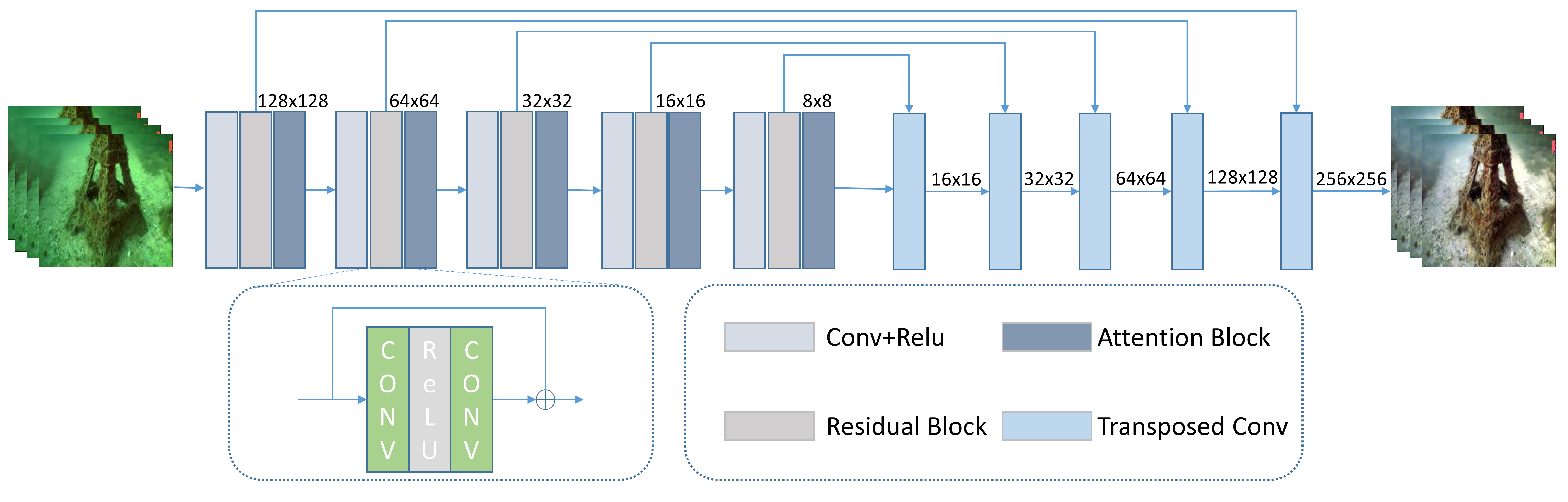}
    \caption{The proposed MuLA-GAN architecture, encompasses an encoder section with convolutional layers, residual blocks, and attention blocks. The decoder section employs straightforward transposed convolution layers. It also provides insight into the internal structure of the residual block, tasked with enhancing features and capturing intricate details.}

    \label{fig:Genrator}
\end{figure*}

\subsection{Deep learning-based image enhancement techniques}

Deep learning techniques have been successfully applied to underwater image enhancement tasks, helping to improve the quality and visibility of images captured in underwater environments. In this section, we briefly discuss a few examples in this regard. For instance, Li et al. \cite{d1}, developed WaterGAN deep learning-based generative adversarial network model that performs network training for reducing the color deviation in underwater images. Fabbri et al. \cite{d2}, developed UGAN by taking the idea of CycleGAN \cite{d3} that performs domain transformation for the reconstruction of degraded images. Guo et al. \cite{d4} developed a novel multi-scale dense GAN by adding a residual multi-scale dense block to the generator resulting in improved enhancement performance. Islam et al. \cite{d5} presented FUnIE-GAN by introducing a multi-model objective function in the network that is specifically designed for online underwater image enhancement. Similarly, Li et al. \cite{d6} worked on the synthesized underwater image enhancement by developing UWCNN model. A gated fusion-based model is also proposed by Li et al. \cite{d7} and named as Water-Net. Islam et al. \cite{d8} introduced a novel Deep SESR model by adding dense residual-in-residual sub-networks that provide both image enhancement and high-resolution underwater images.

Qi et al. \cite{d9} proposed SGUIENet for image enhancement using semantic information as input for region-wise feature learning in the enhancement process. Such design allowed for better extraction of local features for enhancement, resulting in a consistent and superior enhancement. Huang et al. \cite{d10} proposed a novel adaptive group attention (AGA) used to select complementary channels based on dependencies resulting in a reduction in the number of parameters in attention. Furthermore, the proposed AGA is integrated with Swin Transformer for an end-to-end underwater image enhancement network. Cai et al. \cite{d11}, proposed a CURE-Net that systematically enhances image quality stepwise. The proposed model comprises three subnetworks, where the initial two subnetworks employ attention mechanisms and gate fusion to acquire comprehensive contextual features across multiple scales, while the third subnetwork focuses on preserving intricate spatial details. Additionally, a detailed enhancement block and a supervised recovery block were also incorporated between these subnetworks to further enhance color accuracy and refine fine details in the image restoration process. 

\section{Methodology}\label{sec:meth}
To achieve automated image enhancement, our focus lies in acquiring a mapping G: X → Y. This mapping connects a source domain X, which encompasses distorted images, to a desired domain Y, containing their enhanced counterparts. Our chosen approach centers on adopting a conditional GAN-based model as shown in Fig. \ref{fig:Main_Diagram} Within this model, the generator takes on the task of learning this intricate mapping. This learning process unfolds through a dynamic interplay with an adversarial discriminator, set as an iterative min-max game

\subsection{Generator Architecture}
Inspired by the principles of conditional GANs, we developed a deep neural network architecture designed for automatic image enhancement tasks, specifically tailored to improve image quality from a source domain (distorted images) to a desired domain (enhanced images). As shown in Fig. \ref{fig:Genrator}. The generator architecture leverages the foundation of a U-Net-based structure, renowned for its effectiveness in image-to-image translation problems while emphasizing efficiency for fast inference.

The network takes as input images with dimensions of 256x256 pixels and 3 color channels. It follows a fully convolutional design, with each layer applying 2D convolutions using 4x4 filters. Batch Normalization and Leaky-ReLU activation functions are used after each convolution layer to facilitate network training and stability.
The encoder consists of 5 blocks which progressively reduce the spatial dimensions and learn features. It starts with a convolutional layer with 32 output channels and strides of 2, followed by a residual block to capture and enhance image details. Subsequent layers (down2-down5) employ similar structures with an increasing number of output channels, encouraging the network to learn complex representations.
To enhance the network's capacity for capturing fine-grained features and context, Spatial and Channel Attention (SCA) modules are integrated into the architecture. These modules act as mechanisms for focusing on relevant spatial regions and channel-wise information, respectively. These attention mechanisms are integrated at multiple stages in the network to enrich the feature representations.
The decoder upscales the feature maps and fuses information from the encoder using skip connections. Each decoder block consists of a transposed convolutional layer to increase spatial resolution, followed by batch normalization and ReLU activation. The final stage of the network involves upsampling the feature maps, followed by a convolutional layer with 4x4 filters to produce the enhanced output image with the same dimensions as the input.

\subsubsection{Spatio-Channel Attention}
\begin{figure*}[t]
    \centering
    \includegraphics[width=\textwidth]{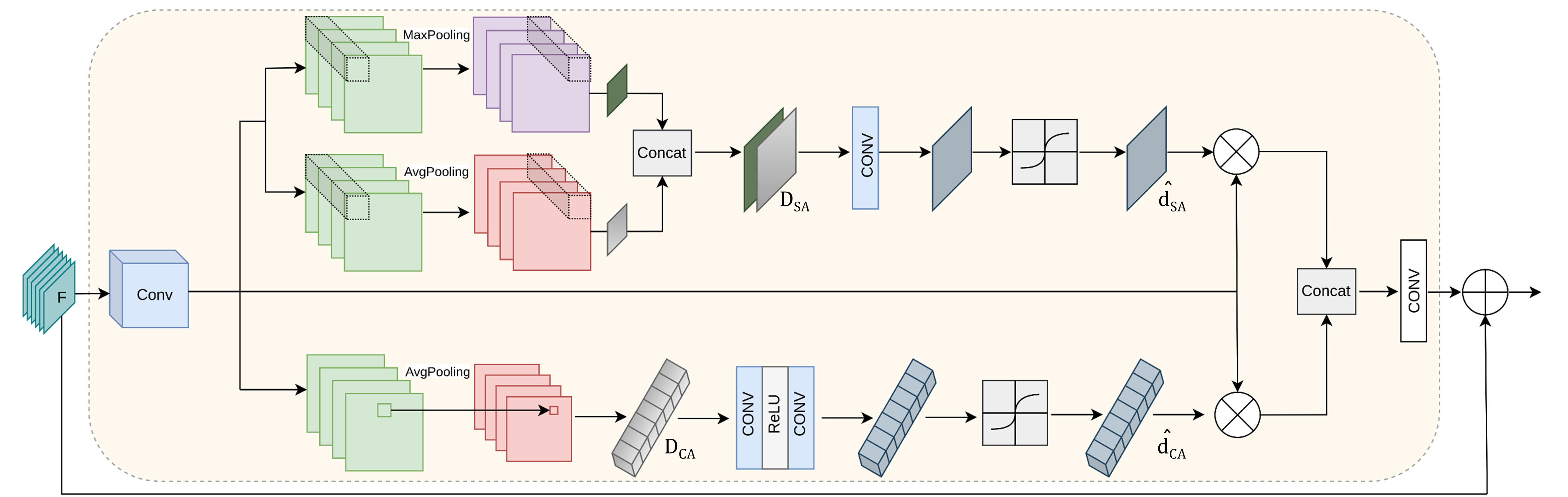}
    \caption{Spatio-Channel attention module incorporating both channel domain attention and spatial domain attention.}
    \label{fig:SCA}
\end{figure*}

Inspired by recent advancements in low-level vision techniques, which rely on attention mechanisms \cite{zhang2018image,zhang2019residual,zamir2022,s&e,CBAM}, we utilized the twofold attention unit named spatio-channel attention (SCA) to extract features in the convolutional feature maps over multiple streams of information. The schematic of spatio-channel attention is shown in Fig.\ref{fig:SCA}.

The SCA efficiently diminishes less pertinent features, allowing only the transmission of more informative ones to the next layers. This feature recalibration is done through the utilization of channel attention \cite{s&e} and spatial attention \cite{CBAM} mechanisms.

A channel attention map is generated by exploiting the inter-channel relationships within feature maps. Each channel of a feature map is treated as a feature detector. The primary objective of channel attention is to identify and emphasize the most meaningful information within an input image.
To efficiently compute channel attention, a two-step process is proposed. First, a "squeeze" operation is applied to the input feature map, which involves global average pooling across the spatial dimensions. This step is designed to capture global context information. As a result of this operation, a feature descriptor denoted as $D_{CA}$ is obtained. $D_{CA}$ is a tensor with dimensions $\mathbb{R}^{1 \times 1 \times C}$, where $C$ represents the number of channels in the feature map $F$ with spatial dimensions $H \times W$. This feature descriptor encapsulates important channel-wise information that can be used to enhance the feature representation.

Subsequently, the "excitation" operator is applied to $D_{CA}$, by passing it through a couple of convolutional layers, followed by a sigmoid gating function. This process generates activation values denoted as $\hat{d_{CA}}$ with dimensions $\mathbb{R}^{1 \times 1 \times C}$.
Finally, the output of the channel attention (CA) branch is obtained by scaling the original channel attention map $F$ with the activation values $\hat{d}$. This scaling operation enhances the relevance of different channels in the feature map, improving the representation of important features in the network.

A spatial attention map is produced by exploiting the inter-spatial relationships among features. Unlike channel attention, which emphasizes 'what' is meaningful within an input, spatial attention is concerned with 'where' the informative regions are located, offering a complementary perspective.
To create the spatial attention map, the SA branch begins by separately applying global average pooling and max pooling operations along the channel dimensions of the feature map $F$. The results of these pooling operations are concatenated to create a feature descriptor represented as $D_{SA}$ with dimensions $\mathbb{R}^{H \times W \times 2}$. Subsequently, this feature descriptor undergoes a convolution operation followed by a sigmoid activation function. This process yields the spatial attention map, denoted as $\hat{d_{SA}}$, which has dimensions $\mathbb{R}^{H \times W \times 1}$. This spatial attention map $\hat{d_{SA}}$ is then multiplied to rescale the original feature set $F$, contributing to the refinement of feature representations within the neural network.

\subsection{Disciminator Architecture}
For the discriminator, we adopted the structural paradigm of Markovian PatchGANs \cite{PatchGAN}. This structural approach utilizes a patchwise processing of the input image, yielding a characteristic matrix of dimensions $n × n × 1$. Each element within this matrix corresponds to a distinct patch in the input image, thus enabling the discrimination of the entire image by assessing all constituent patches. Notably, this architectural choice rests upon the assumption of pixel independence beyond the patch size, implying that the discrimination of images depends solely on patch-level information. This assumption bears particular significance in effectively capturing high-frequency features, such as local textures and styles \cite{yi2017dualgan}. Furthermore, this structural design offers computational efficiency benefits by minimizing the number of parameters through patch-level comparisons. As depicted in Fig, the discriminator takes both the generated image and the ground truth as input for identification purposes. Following this, a sequence of five convolutional layers, coupled with four nonlinear activation functions, is utilized to render a binary output of real or fake. The initial four layers of the discriminator employ a 3 × 3 convolutional kernel with a stride of 2, while the final layer adopts a 4 × 4 convolutional kernel with a stride of 1. Rectified Linear Unit (ReLU) activation functions are applied to introduce nonlinearity into the network architecture.

\subsection{Loss Functions}
In this work, we used a linear combination of the  adversarial loss fucntion $L_adv$ , $L_1$ loss function and the perceptual loss function $L_{\text{per}}$.

\begin{equation}
 L_{\text{total}} = L_{\text{adv}} + \lambda_1 L_{\text{per}} + \lambda_2 L_1 
\end{equation}

\noindent
where, $\lambda_1$ and $\lambda_2$ are scaling factors.

\textbf{Adversarial Loss:} We used adversarial loss to facilitate the training of the generator network and the discriminator network in a competitive manner. This loss function encourages the generator to minimize the log probability that the discriminator assigns to the generated samples, effectively pushing the generator to produce more enhanced images similar to reference images.
Mathematically, it can be expressed as:
\begin{equation}
  L_{\text{adv}}(G, D) = \mathbb{E}_{X, Y} \left[\log D(X, Y)\right] + \mathbb{E}_{X, Z} \left[\log(1 - D(X, G(X,Z)))\right]     
\end{equation}

\noindent
where X represents the input images, Y represents the reference images, and G(X,Z) represents the generated images.

\textbf{Perceptual loss:} We also used a perceptual loss inspired by the work of Johnson et al \cite{Perceptual}, aiming to capture the global dependency between underwater images and ground truth images by considering deep features. This perceptual loss is defined based on the VGG-19 network \cite{VGG19}, which is a deep convolutional neural network pre-trained on a large dataset. The perceptual loss, denoted as $L_{\text{per}}$, quantifies the dissimilarity between feature representations of the generated underwater image and the ground truth image as captured by VGG-19. It is calculated as the sum of the $\ell_1$ absolute difference between feature maps in a chosen layer of the VGG-19 network. Mathematically, the perceptual loss can be expressed as:
\begin{equation}
  L_{\text{per}} = \frac{1}{C_j H_j W_j} \sum_{j} \left\|\Phi_j({Y}) - \Phi_j(G(X,Z))\right\| 
\end{equation}

Where, $L_{\text{per}}$ represents the perceptual loss.
$j$ refers to the index of the chosen layer in the VGG-19 network.
$C_j$, $H_j$, and $W_j$ are the channel number, height, and width of the feature map in the $j$th layer of VGG-19.
$\Phi_j({G(X,Y)})$ is the feature representation of the generated underwater image $G(X,Y)$ in the $j$th layer.
$\Phi_j(Y)$ is the feature representation of the ground truth image $Y$ in the $j$th layer.
In this formulation, the perceptual loss measures the dissimilarity between feature maps extracted from the VGG-19 network, helping to guide the optimization process toward generating images that align more closely with the features of the ground truth images.

\textbf{MAE loss:} We also used the MAE loss (L1 loss) to enhance the model's ability to capture fine-grained details while mitigating blurring effects, contributing to improved sample quality and overall performance. L1 loss can be mathematically expressed as:
\begin{equation}
 L_1(G) = \mathbb{E}_{X,Y,Z} \left[ \left\|Y - G(X, Z)\right\| \right]   
\end{equation}

Where Y represents the reference image in our target domain and $G(X,Z)$ represents the generated image from the generator.

\section{Experimentation}\label{sec:exp}
\noindent \textbf{Datasets:} 
In this study, we experimented with the Underwater Image Enhancement Benchmark (UIEB) dataset for both training and testing our model. The UIEB dataset consists of 950 real-world underwater images, out of which 890 images have corresponding reference images. These original images were captured from various real-world underwater environments. To create the reference images, 12 different enhancement methods were applied to generate candidate images, and multiple volunteers participated in voting for each pair of enhanced images. The image with the highest number of votes was selected as the corresponding reference image\cite{d7}. The UIEB dataset provides an excellent platform for evaluating diverse underwater image enhancement algorithms, considering its rich and varied underwater scenes, along with images exhibiting varying degrees of degradation. For our experimentation, we randomly chose 800 pairs of real-world images for training. The test dataset comprises the remaining 90 real-world images in the UIEB dataset, 60 challenge images (no reference), and U45 \cite{li2019fusion}, the underwater color cast set (UCCS) dataset \cite{8949763}.

\noindent \textbf{Experiment Settings:} 

To ensure a comprehensive evaluation of our model, we conducted qualitative and quantitative comparison experiments with several state-of-the-art methods. To maintain consistency, we utilized the source codes provided by the respective authors or referred to the descriptions in the original papers. Additionally, for some models, we relied on open-source resources to construct the necessary models for evaluation. Throughout the experiments, all the methods underwent the same testing and training processes. The evaluation environment was set up with an NVIDIA RTX 6000 48GB GPU, Intel Xeon(R) Gold 6230 CPU @ 2.10GHz × 80 to ensure fair and comparable results.

\subsection{Evaluation Metrics:}
In our experiments, we rigorously evaluated multiple models using widely known underwater image quality evaluation metrics. We evaluated the models extensively for part of UIEB test dataset which contains reference images using two core metrics: Peak Signal-to-Noise Ratio (PSNR) \cite{PSNR} and Structural Similarity Index (SSIM) \cite{SSIM}. These metrics are critical in assessing the degree of similarity between the image content and structural texture and the reference image. Higher PSNR and SSIM values indicate more fidelity, implying a closer alignment with the reference image.

For datasets with no reference images, we utilized specific metrics for underwater images. These metrics include underwater image quality measures (UIQM)\cite{UIQM}, underwater color image quality evaluation (UCIQE) \cite{UCIQE}, and naturalness image quality evaluator (NIQE) \cite{NIQE}. Through these metrics, we sought to capture crucial aspects related to the consistency of the enhanced images with human visual perception, the appropriate balance of hue, saturation, and contrast, and the overall visual quality. Higher UIQM and UCIQE scores are indicative of closeness to human visual perception, while a lower NIQE score suggests an elevated level of visual quality in the image.
UCIQE is a composite measure based on chroma $\sigma_c$, contrast $con_l$, and saturation $\mu_s$, computed as follows:

\begin{figure*}
    \centering
    \includegraphics[width=\textwidth]{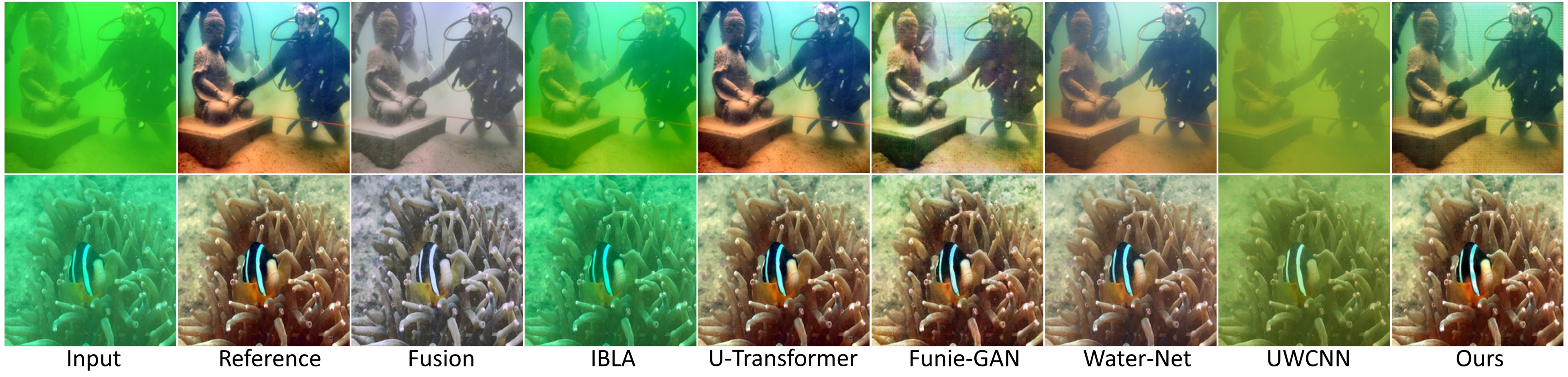}
    \caption{A comparative visual analysis of SOTA Methods for greenish color scenes. Left to right columns show, Input image, Reference image, and outputs of different methods such as Fusion\cite{Fusion}, IBLA\cite{IBLA}, U-Transformer\cite{U-Trans}, Funie-GAN\cite{d5}, Water-Net\cite{d7}, UWCNN\cite{d6} and Ours}
    \label{fig:example1}
\end{figure*}

\begin{figure*}
    \centering
    \includegraphics[width=\textwidth]{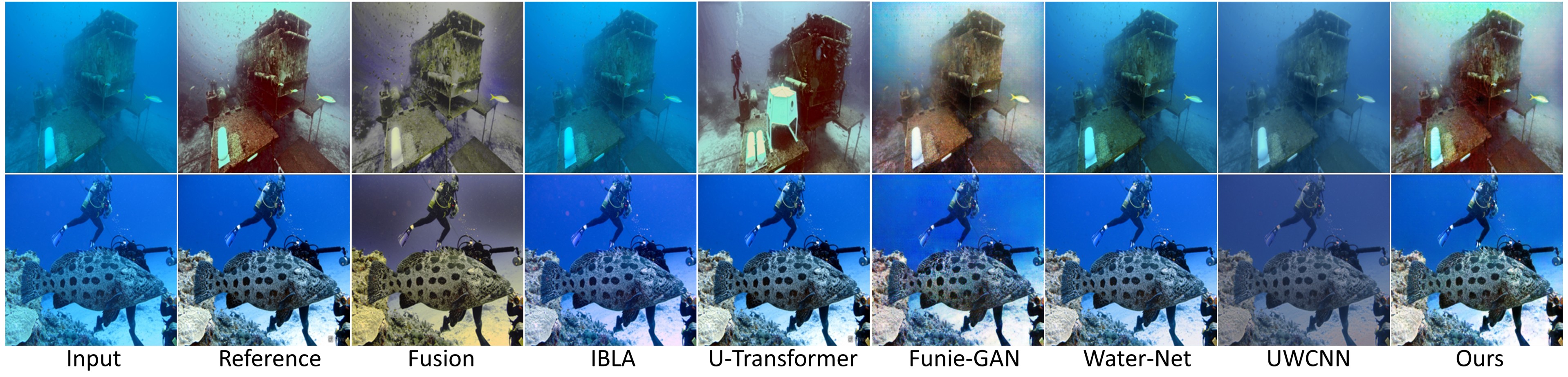}
    \caption{A comparative visual analysis of SOTA Methods for blueish color scenes.Left to right columns show, Input image, Reference image, and outputs of different methods such as Fusion\cite{Fusion}, IBLA\cite{IBLA}, U-Transformer\cite{U-Trans}, Funie-GAN\cite{d5}, Water-Net\cite{d7}, UWCNN\cite{d6} and Ours}
    \label{fig:example2}
\end{figure*}

\begin{figure*}
    \centering
    \includegraphics[width=\textwidth]{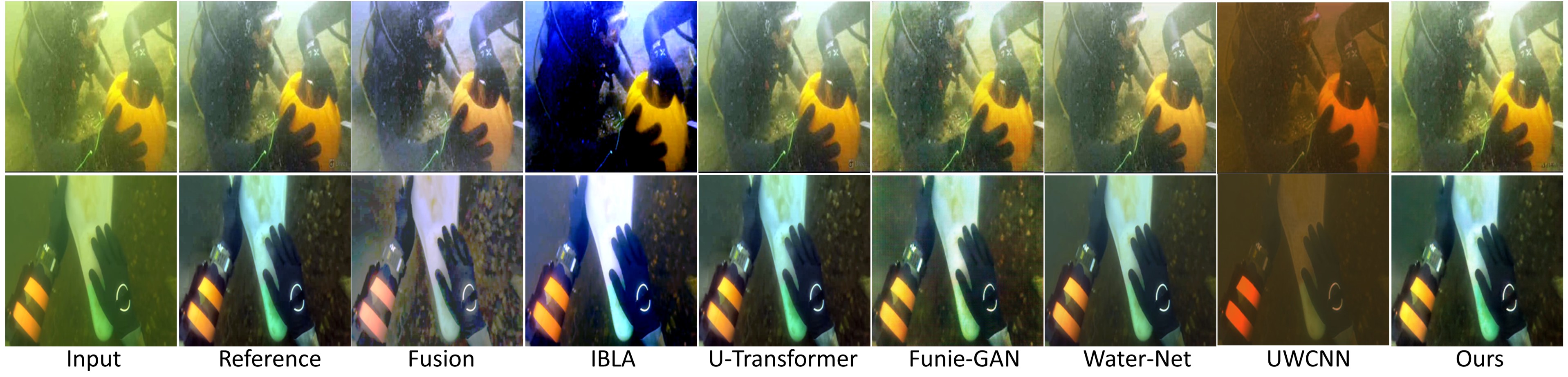}
    \caption{A comparative visual analysis of SOTA Methods for yellowish color scenes..Left to right columns show, Input image, Reference image, and outputs of different methods such as Fusion\cite{Fusion}, IBLA\cite{IBLA}, U-Transformer\cite{U-Trans}, Funie-GAN\cite{d5}, Water-Net\cite{d7}, UWCNN\cite{d6} and Ours}
    \label{fig:example3}
\end{figure*}

\begin{equation}
UCIQE = c_1 \cdot \sigma_c + c_2 \cdot con_l + c_3 \cdot \mu_s
\end{equation}
where the coefficients are $c_1 = 0.4680$, $c_2 = 0.2745$, and $c_3 = 0.2575$, as defined in \cite{UCIQE}. A higher UCIQE score indicates an improved balance between chroma, contrast, and saturation.
Meanwhile, UIQM is formed through a linear combination of UICM, UISM, and UIConM, representing colorfulness, sharpness, and contrast metrics, respectively:
\begin{equation}
UIQM = c_1 \cdot UICM + c_2 \cdot UISM + c_3 \cdot UIConM
\end{equation}
with coefficients $c_1 = 0.0282$, $c_2 = 0.2953$, and $c_3 = 3.5753$, as suggested in \cite{UIQM}. A higher UIQM score reflects an enhanced trade-off among colorfulness, sharpness, and contrast in the evaluated results.

\section{QUALITATIVE ANALYSIS}
The UIEB, a dataset containing a diverse range of degraded underwater images, serves as a common benchmark for evaluating underwater image enhancement techniques. Initially, a selection of underwater images from the UIEB was made based on the visual appearance, and they were categorized into eight appearance-based groups: greenish, bluish, yellowish, downward-looking, upward-looking, forward-looking, low backscattered, and high backscattered.

Within open-water environments, the sequence of light attenuation is distinctive, with red light being the first to diminish due to its longer wavelength, followed by green light, blue light and lastly yellow light \cite{8099551}. This selective attenuation characteristic of open water often gives rise to underwater images exhibiting a yellowish, bluish, or greenish cast, such as those exhibited in Figure 4-5-6. The resultant color deviation significantly undermines the visual fidelity of these underwater images, presenting a challenge in terms of rectification.

As depicted in Figure 4, IBLA did not effectively eliminate the greenish color, while Fusion resulted in a somewhat grayish tone replacing the green. Among the deep learning techniques, Funie-GAN successfully removed the green hue but introduced some flickering noise into the images. On the other hand, UWCNN was not able to significantly enhance the images. Water-Net, U-Transformer and our method, however, managed to successfully eliminate the greenish tint and improve image quality. While there may be some minor color differences, overall, the image quality has been notably enhanced. 

Figure 5 illustrates a visual comparison of bluish color correction. In this context, IBLA did not significantly contribute to image enhancement, and Fusion introduced a form of haze. Among the deep learning-based methods, UWCNN performed the poorest. Water-Net retained some bluish content in the resulting images, whereas Funie-GAN, U-Transformer, and our approach successfully enhanced the images and effectively corrected the color.

In Figure 6, observations reveal that IBLA has introduced a bluish tint in a few images. Fusion has struggled to enhance contrast in the image. Among the deep learning-based methods, Water-Net introduced a bluish tint in one image, and UWCNN has darkened the overall visual appearance. Funie-GAN still exhibits some residual yellowish artifacts, while U-Transformer and our method has effectively eliminated the yellowish haze.

Hence, when it comes to color correction, our method has consistently outperformed other approaches. While various methods exhibited strengths and weaknesses in addressing color issues, our method consistently delivered superior results, making it the most effective choice for color correction.
\begin{figure*}
    \centering
    \includegraphics[width=\textwidth]{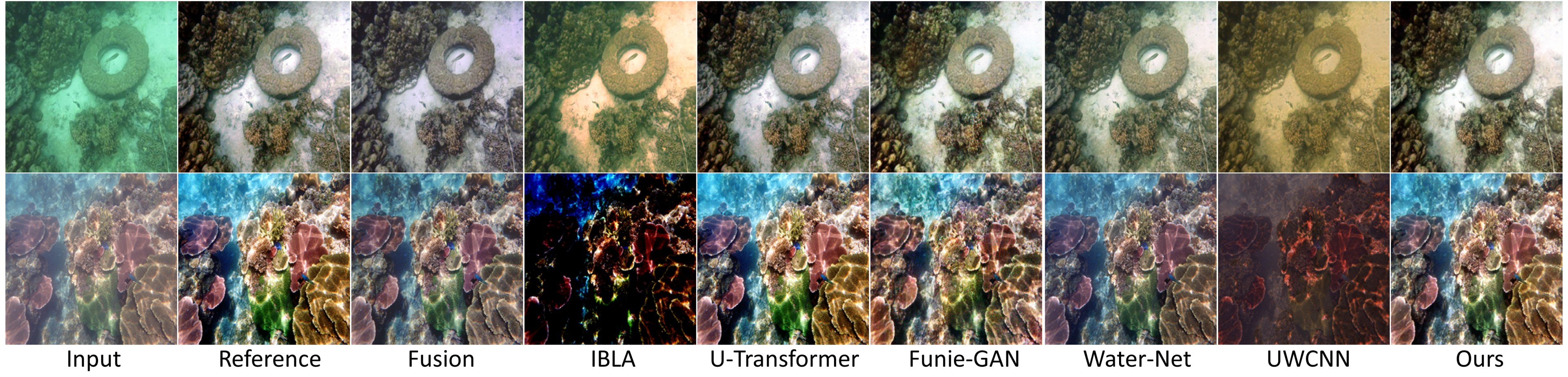}
    \caption{A comparative visual analysis of SOTA Methods for downward-looking scenes.Left to right columns show, Input image, Reference image, and outputs of different methods such as Fusion\cite{Fusion}, IBLA\cite{IBLA}, U-Transformer\cite{U-Trans}, Funie-GAN\cite{d5}, Water-Net\cite{d7}, UWCNN\cite{d6} and Ours}
    \label{fig:example4}
\end{figure*}

\begin{figure*}
    \centering
    \includegraphics[width=\textwidth]{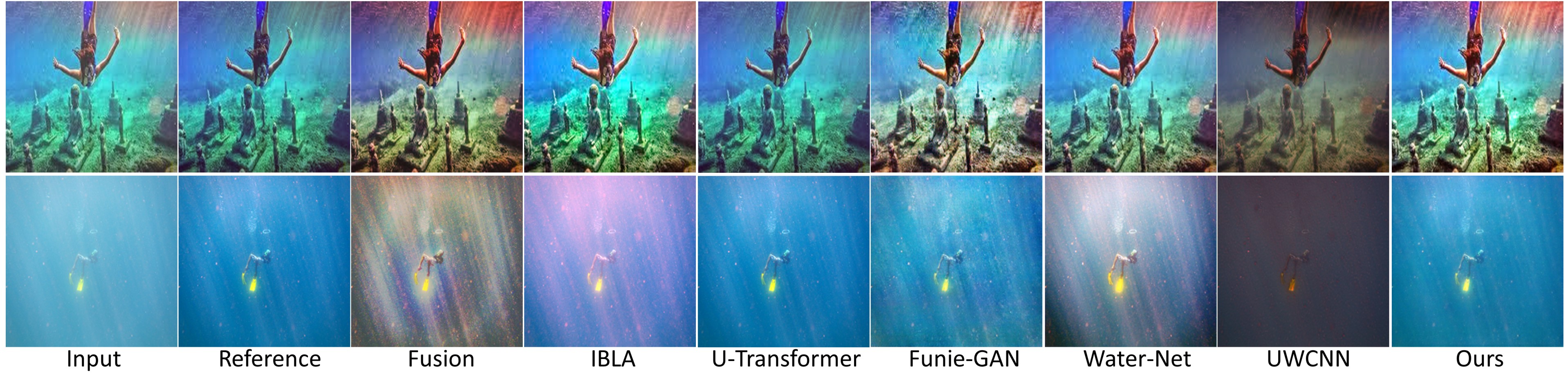}
    \caption{A comparative visual analysis of SOTA methods for upward-looking scenes. Left to right columns show, Input image, Reference image, and outputs of different methods such as Fusion\cite{Fusion}, IBLA\cite{IBLA}, U-Transformer\cite{U-Trans}, Funie-GAN\cite{d5}, Water-Net\cite{d7}, UWCNN\cite{d6} and Ours}
    \label{fig:example5}
\end{figure*}

\begin{figure*}
    \centering
    \includegraphics[width=\textwidth]{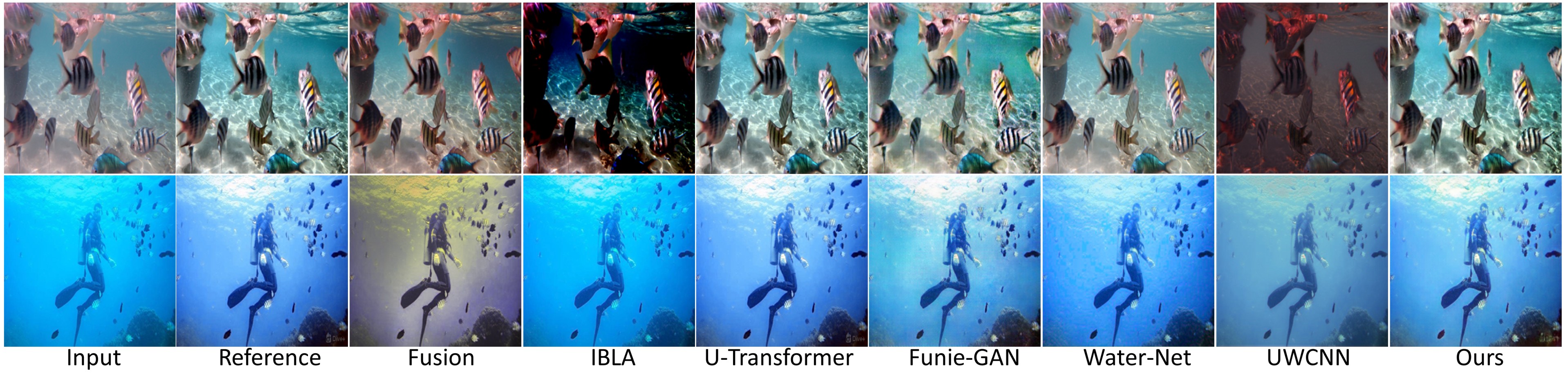}
    \caption{A comparative visual analysis of SOTA Methods for forward-looking scenes. Left to right columns show, Input image, Reference image, and outputs of different methods such as Fusion\cite{Fusion}, IBLA\cite{IBLA}, U-Transformer\cite{U-Trans}, Funie-GAN\cite{d5}, Water-Net\cite{d7}, UWCNN\cite{d6} and Ours}
    \label{fig:example6}
\end{figure*}
In the context of scene perspective from various angles, downward-looking, upward-looking, and forward-looking images are illustrated in Figures 7, 8, and 9.

In Figure 7, Fusion has impressively mitigated the impact of haze on underwater images. Notably, methods such as IBLA, and UWCNN tended to introduce color distortions in their enhanced outputs. While Water-Net still retains some remnants of the hazy effect, other methods have shown significant improvements in haze removal.

Moreover, in Figure 8, methods such as Fusion, IBLA, and UWCNN faced difficulties in accurately estimating veiling light from upward-looking images. These challenges extended to inaccuracies in estimating veiling light from RGB values, especially in textured regions of upward-looking images. While Funie-GAN, U-Transformer and Water-Net exhibited some remaining color deviations, Funie-GAN also introduced blurriness. Overall, our method has demonstrated superior performance compared to all other methods. 

In forward-looking images, as depicted in Figure 9, it's worth noting that, except for IBLA and UWCNN, all other methods have shown significant improvements in enhancing image quality. Overall, our proposed method demonstrated exceptional performance by achieving optimal exposure and intricate detail restoration across all three angles.

\begin{figure*}[t]
    \centering
    \includegraphics[width=\textwidth]{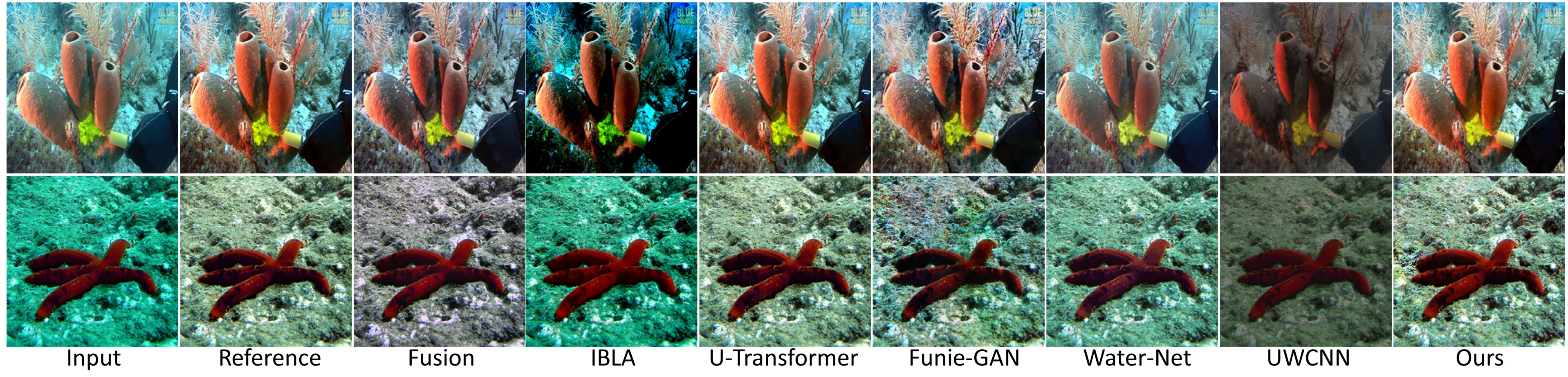}
    \caption{A comparative visual analysis of SOTA Methods for low-back scattered scenes.Left to right columns show, Input image, Reference image, and outputs of different methods such as Fusion\cite{Fusion}, IBLA\cite{IBLA}, U-Transformer\cite{U-Trans}, Funie-GAN\cite{d5}, Water-Net\cite{d7}, UWCNN\cite{d6} and Ours}
    \label{fig:example7}
\end{figure*}

\begin{figure*}[t]
    \centering
    \includegraphics[width=\textwidth]{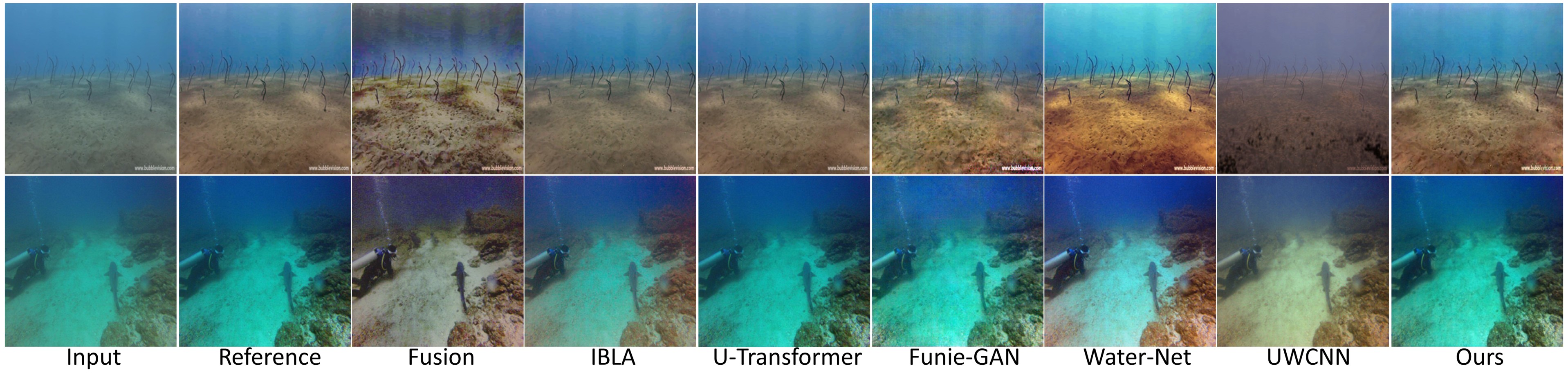}
    \caption{A comparative visual analysis of SOTA Methods for high backscattered scenes. Left to right columns show, Input image, Reference image, and outputs of different methods such as Fusion\cite{Fusion}, IBLA\cite{IBLA}, U-Transformer\cite{U-Trans}, Funie-GAN\cite{d5}, Water-Net\cite{d7}, UWCNN\cite{d6} and Ours}
    \label{fig:example8}
\end{figure*}

\begin{table*}[t]
\caption{Full-reference image quantitative comparison on UIEB test dataset by using PSNR and SSIM.}
\centering
\label{tab:Full_reference}
\begin{tabular}{|l|l|l|l|l|l|l|l|}
\hline
Method     & Fusion \cite{Fusion}    & IBLA \cite{IBLA}      & U-Transformer \cite{U-Trans}& Funie-GAN \cite{d5} & Water-Net \cite{d7} & UWCNN \cite{d6}     & Ours      \\ \hline
PSNR       & 20.87     & 19.04     & 21.64& 21.03     & \underline{24.36}     & 15.25     & \textbf{25.59}     \\ \hline
SSIM       & 0.857     & 0.703     & 0.778& 0.775     & \underline{0.885}     & 0.656     & \textbf{0.893}     \\ \hline
\end{tabular}
\end{table*}

Scattering induces intricate alterations in the trajectory of light propagation, thereby influencing the spatial and temporal distribution of light energy. Within underwater images, the scattering phenomenon is contingent on the distance between the scene and the camera. As evident in Figure 10, mitigating the effects of fog is relatively straightforward in low-scatter scenes characterized by short distances. However, in such scenes, IBLA and UWCNN have introduced unwanted darkness, while Funie-GAN introduced blurriness. On the other hand, Fusion, Water-Net, U-Transformer and our method have excelled in enhancing image quality under these conditions.

Conversely, Figure 11 illustrates the difficulty in rectifying blur effects prevalent in high-scatter scenes that entail greater distances.
The challenge becomes apparent in the context of highly scattered scenes, as inaccuracies in models hinder the capacity of physical model-based methods (such as Fusion and IBLA) to eliminate scattering, as observed in Figure 11. In contrast, Funei-GAN, water-net, U-Transformer, and our technique exhibit superior performance in addressing this issue. Under conditions of extreme degradation, characterized by weak illumination and substantial scattering of suspended particles, underwater images contend with issues like color distortion, reduced contrast, and blurriness. These combined factors pose a considerable challenge in the realm of underwater image enhancement. Despite these formidable challenges, our model has consistently excelled in enhancing image quality.

\section{Quantitative Analysis}
To quantitatively evaluate the performance of different methods, we conduct both full-reference and non-reference evaluations.

\subsection{Full-reference Evaluation}
In the evaluation of the test datasets containing reference images, a full-reference assessment is conducted utilizing two commonly-utilized metrics: PSNR (Peak Signal-to-Noise Ratio), SSIM (Structural Similarity Index. These metrics enable an objective evaluation of the method's performance. Higher PSNR and SSIM scores indicate a closer resemblance of the result to the reference image.

As shown in Table 1. IBLA achieved moderate results while Fuison performed comparatively better with 20.87, and 0.857 and for PSNR score and SSIM, respectively, indicating good performance in noise reduction and structural preservation.
While UWCNN faced challenges in image enhancement, with a PSNR of 15.25 and an SSIM of 0.656, indicating limited noise reduction and structural preservation, other deep learning-based methods have performed quite well. U-Transformer and Funie-GAN delivered better results with a commendable PSNR of 21.64 and 21.03 respectively, and an SSIM of 0.778 and 0.775 respectively, signifying reduced distortion and substantial structural similarity. Water-Net and Ours exhibited outstanding results. Our method outperformed Water-Net, achieving a higher PSNR of 25.59 compared to Water-Net's 24.36, and achieving the highest SSIM value of 0.893 compared to Water-Net's 0.885.

\subsection{Non-Reference Evaluation}

In the evaluation of test datasets that do not contain reference images, three non-reference metrics, namely UCIQE \cite{UCIQE} and UIQM \cite{UIQM}, and NIQE \cite{NIQE} are used. A higher score in either UCIQE or UIQM signifies an enhanced level of human visual perception in the results. while a lesser score in NIQE indicates a better perception quality.
In the evaluation of the C-60 challenge dataset, our method demonstrated remarkable performance across various quality metrics. As shown in Table 2. our method achieved the highest UIQM score of 3.12, surpassing all other approaches. Funie-GAN achieved the second-highest UIQM score of 3.10. When considering a balance between chroma, contrast, and saturation, our method once again excelled with the highest UCIQE score of 0.591, followed closely by Water-Net with a score of 0.578. In terms of perception quality, our method outperformed all others, achieving a NIQE value of 4.67, while Funie-GAN attained a value of 4.73.

Table 3. shows the assessment of the U45 dataset, our method attained a UIQM score of 3.23 and a UCIQE score of 0.612, making it the top performer in terms of image quality enhancement. Funie-GAN emerged as the second-best method with respective scores of 3.21 for UIQM and 0.602 for UCIQE. Moreover, when evaluating the perception quality, our method demonstrated its superiority once more, achieving a NIQE score of 3.86. Water-Net closely followed with a NIQE score of 3.91.
As shown in Table 5, our method achieved a UIQM score of 3.17 and a UCIQE score of 0.568, showcasing its effectiveness in enhancing image quality for UCCS dataset. Water-Net also demonstrated strong performance with a UIQM score of 3.13, while Funie-GAN achieved a competitive UCIQE score of 0.558. In terms of perception quality, the U-Transformer received an NIQE score of 3.96, with our method closely following with a score of 4.03.

\begin{table}[t]
\caption{Quantitative comparison on C-60 dataset using UIQM, UCIQE, and NIQE metrics.}
{%
\begin{tabular}{|l|l|l|l|}
\hline
          & UIQM$\uparrow$ & UCIQE$\uparrow$ & NIQE$\downarrow$ \\ \hline
Raw\_Images & 1.99 & 0.478 & 5.18 \\ \hline
Fusion    & 2.78 & 0.512 & 4.74 \\ \hline
IBLA      & 1.81 & 0.574 & 5.02 \\ \hline
U-Transformer& 2.65 & 0.534 & 4.94\\ \hline
Funie-GAN & \underline{3.10} & 0.572 & \underline{4.73} \\ \hline
Water-Net & 2.57 & \underline{0.578} & 4.88 \\ \hline
UWCNN     & 2.25 & 0.466 & 4.89 \\ \hline
Ours      & \textbf{3.12} & \textbf{0.591} & \textbf{4.67} \\ \hline
\end{tabular}%
}
\end{table}

\begin{table}[t]
\caption{Quantitative comparison on U45 dataset using UIQM, UCIQE, and NIQE metrics.}
{%
\begin{tabular}{|l|l|l|l|}
\hline
          & UIQM$\uparrow$ & UCIQE$\uparrow$ & NIQE$\downarrow$ \\ \hline
Raw\_Images & 2.44 & 0.481 & 5.03 \\ \hline
Fusion    & 3.14 & 0.532 & 3.97 \\ \hline
IBLA      & 1.60 & 0.579 & 5.11 \\ \hline
U-Transformer& 3.10 & 0.553 & 4.30\\ \hline
Funie-GAN & \underline{3.21} & \underline{0.602} & 3.95 \\ \hline
Water-Net & 3.18 & 0.587 & \underline{3.91} \\ \hline
UWCNN     & 2.82 & 0.471 & 4.52 \\ \hline
Ours      & \textbf{3.23} & \textbf{0.612} & \textbf{3.86} \\ \hline
\end{tabular}%
}
\end{table}

\begin{table}[t]
\caption{Quantitative comparison on UCCS dataset using UIQM, UCIQE, and NIQE metrics.}
{%
\begin{tabular}{|l|l|l|l|}
\hline
          & UIQM$\uparrow$ & UCIQE$\uparrow$ & NIQE$\downarrow$ \\ \hline
Raw\_Images & 2.29 & 0.410 & 4.57 \\ \hline
Fusion    & 2.99 & 0.476 & 4.38 \\ \hline
IBLA      & 2.36 & 0.480 & 4.29 \\ \hline
U-Transformer& 3.02 & 0.539 & \textbf{3.96}\\ \hline
Funie-GAN & 3.05 & \underline{0.558} & 4.37 \\ \hline
Water-Net & \underline{3.13} & 0.550 & 4.07 \\ \hline
UWCNN     & 2.78 & 0.455 & 4.12 \\ \hline
Ours      & \textbf{3.17} & \textbf{0.568} & \underline{4.03} \\ \hline
\end{tabular}%
}
\end{table}

Overall, deep learning-based methods demonstrated superior performance compared to prior-based methods in enhancing the quality of images in the mentioned datasets. Among these, our method emerged as the top-performing one, showcasing its effectiveness in improving image quality and perceptual fidelity, making it a valuable choice for enhancing images.

\section{Evaluation on locally collected dataset}

\begin{figure*}[t]
    \centering
    \includegraphics[width=\textwidth]{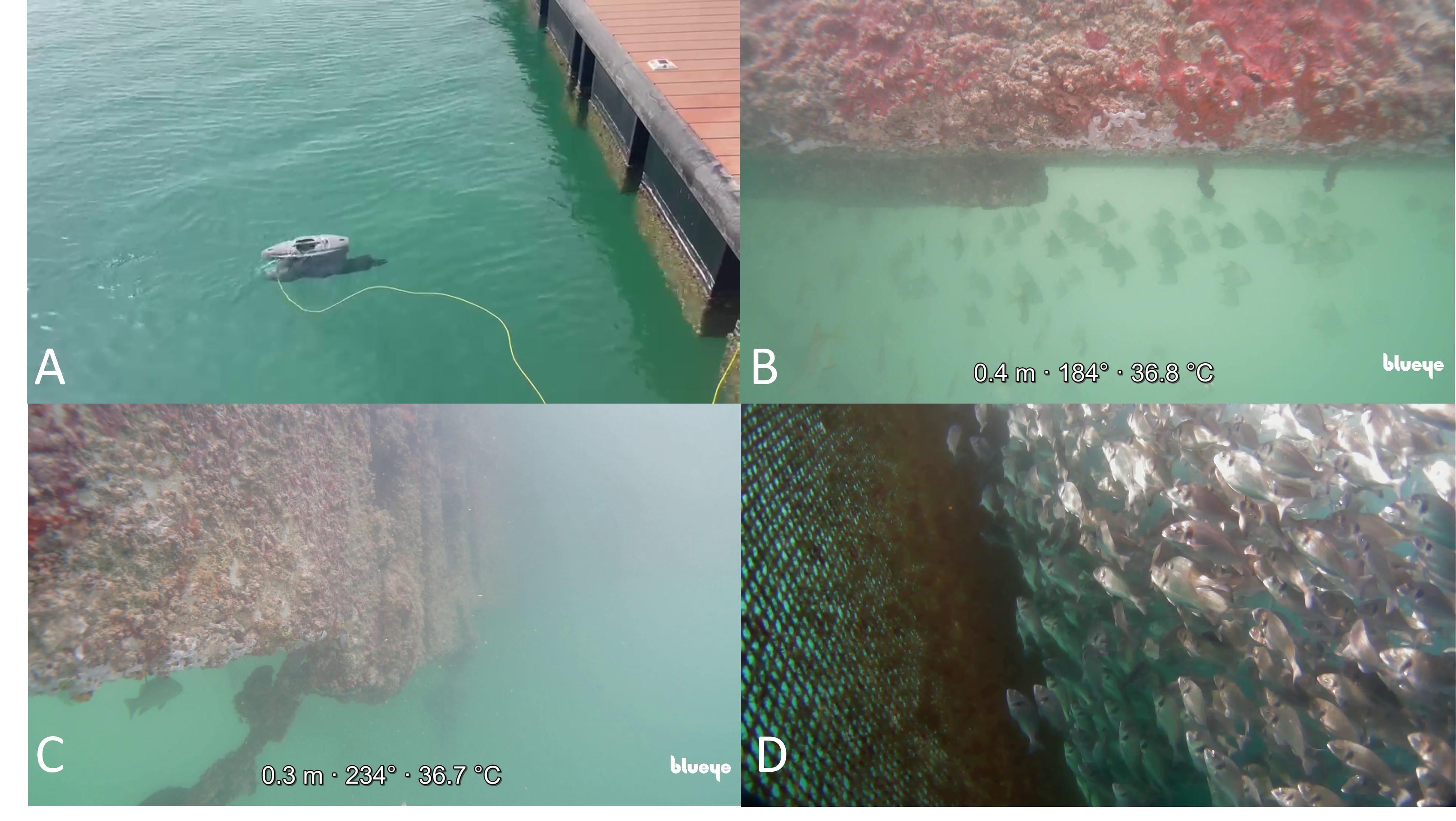}
    \caption{Data collection. (A) shows Blueye ROV used for data collection, (B), (C), and (D) show the different scenes captured during the experiments.}
    \label{fig:datawa}
\end{figure*}

In this study, we also validated our proposed method on our locally collected dataset for underwater bio-fouling detection in an aquaculture environment at Saadiyat Beach, Abu Dhabi. Primarily, the dataset contains images of fish and biofouling grown on the aqua-net surface \cite{akram2024aquaculture,akram2022visual}. These images were acquired using the Blueye ROV as shown in Figure \ref{fig:datawa}. The ROV has an installed HD camera, which allows the recording of high-resolution images of size 1080x1920 pixels. A total of 159 images are selected by getting 1 frame per 2 seconds. This comprehensive evaluation covered a wide range of scenes, encompassing various lighting conditions, scattering effects, and perspective views. As shown in Fig. 12, our method consistently demonstrated excellent performance by effectively removing haze and enhancing images across these diverse scenarios.

On the other hand, among the prior-based methods, Fusion exhibited better performance by mitigating the haze effect, although it fell short of achieving complete image enhancement. In contrast, IBLA introduced noticeable color deviations rather than enhancement, which was less desirable.
Among the deep learning-based methods, Funie-GAN showcased strong performance but tended to introduce noise artifacts into the images. UWCNN, however, struggled as it darkened the resultant images, impacting overall quality. U-Transformer and Water-Net delivered satisfactory results, although some residual haze persisted in the output images.
\begin{figure*}[t]
    \centering
    \includegraphics[width=\textwidth]{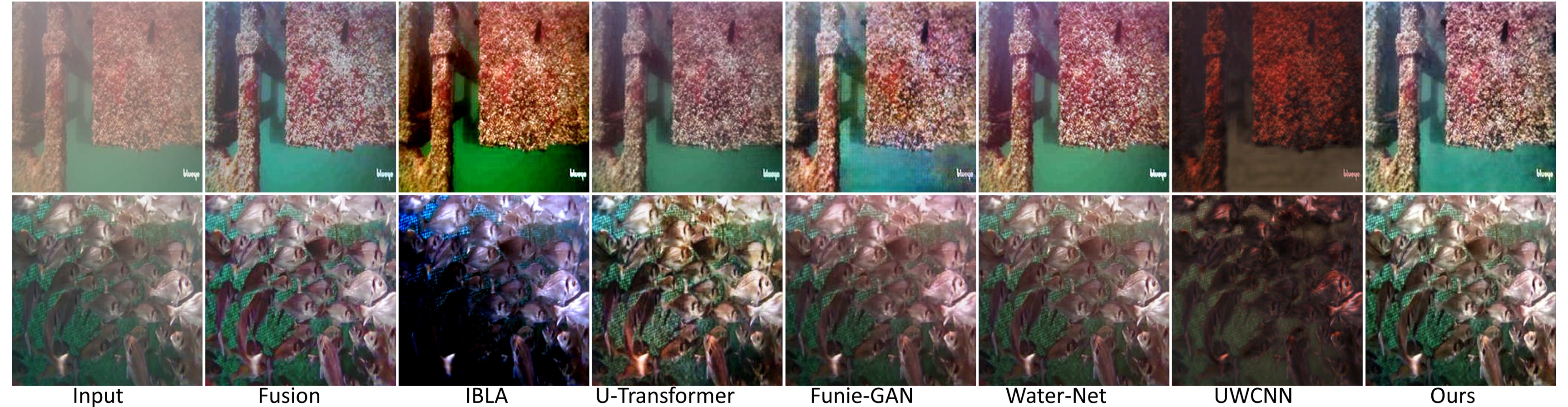}
    \caption{Visual comparison of different methods on the private dataset.Left to right columns show, Input image, Reference image, and outputs of different methods such as Fusion\cite{Fusion}, IBLA\cite{IBLA}, U-Transformer\cite{U-Trans}, Funie-GAN\cite{d5}, Water-Net\cite{d7}, UWCNN\cite{d6} and Ours}
    \label{fig:Unseen}
\end{figure*}

\begin{table}[t]
\caption{Non-reference image quantitative comparison on private data using UIQM, UCIQE, and NIQE.}
\centering
\label{tab:Unseen_data}
\resizebox{6cm}{!}
{
\begin{tabular}{|l|l|l|l|}
\hline
          & UIQM$\uparrow$ & UCIQE$\uparrow$ & NIQE$\downarrow$ \\ \hline
Fusion    & 2.92 & 0.545 & 5.40 \\ \hline
IBLA      & 1.47 & \textbf{0.626} & 6.38 \\ \hline
U-Transformer& 2.99& 0.55& \textbf{5.21}\\ \hline
Funie-GAN & \underline{3.21} & 0.599 & 5.56 \\ \hline
Water-Net & 3.02 & 0.582 & 6.06 \\ \hline
UWCNN     & 2.59 & 0.562 & 5.75 \\ \hline
Ours      & \textbf{3.26} & \underline{0.603}& \underline{5.39} \\ \hline
\end{tabular}%
}
\end{table}

A quantitative analysis of this dataset is given in table 3. confirms the effectiveness of our model. Our method has achieved a UIQM score of 3.26 and an NIQE score of 5.39, followed by the Funie-GAN in terms of UIQM with a score of 3.21 and Fusion in terms of NIQE with a score of 5.40. While measuring UCIQE score, IBLA has achieved a score of 0.626, although their resultant images are just the color deviations from the original images. But since UCIQE measures a linear combination of chroma, saturation, and contrast, a higher value can be expected just by having a change in colors (even if they are not desirable).

\section{Conclusion}
In this study, we introduced an innovative GAN-based architecture designed specifically for underwater image enhancement applications. A key advancement is the incorporation of spatial and channel attention within the generator's encoder, contributing to enhanced feature representation learning and subsequent improvements in image enhancement performance.
Our comprehensive experiments, spanning diverse datasets including the UIEB test dataset, UIEB challenge dataset, U45, and UCCS dataset, demonstrated the superior performance of our model both qualitatively and quantitatively. Notably, the proposed model exhibited a significant performance boost, surpassing Water-Net by 5$\%$ in PSNR value and achieving more than a 1$\%$ improvement in SSIM for the reference image test dataset.
Furthermore, the validation of our trained model on a private underwater imaging dataset acquired from an open sea environment under various conditions showcased impressive performance. This underscores the substantial potential of our model for real-time deployment on underwater vehicles.
In conclusion, our work contributes to the field of underwater image enhancement by introducing a novel GAN-based model with integrated spatial and channel attention mechanisms. The demonstrated improvements across controlled datasets and real-world conditions underscore the robustness and practical applicability of our proposed model, marking a noteworthy advancement in underwater imaging technologies.

\section*{Acknowledgement}
\noindent This work is supported by the Khalifa University of Science and Technology under Award No. MBZIRC-8434000194, CIRA-2021-085, FSU-2021-019, RC1-2018-KUCARS, 8434000534.\\

\bibliographystyle{IEEEtran}
\bibliography{references}

\begin{thebibliography}{10}
\providecommand{\url}[1]{#1}
\csname url@samestyle\endcsname
\providecommand{\newblock}{\relax}
\providecommand{\bibinfo}[2]{#2}
\providecommand{\BIBentrySTDinterwordspacing}{\spaceskip=0pt\relax}
\providecommand{\BIBentryALTinterwordstretchfactor}{4}
\providecommand{\BIBentryALTinterwordspacing}{\spaceskip=\fontdimen2\font plus
\BIBentryALTinterwordstretchfactor\fontdimen3\font minus \fontdimen4\font\relax}
\providecommand{\BIBforeignlanguage}[2]{{%
\expandafter\ifx\csname l@#1\endcsname\relax
\typeout{** WARNING: IEEEtran.bst: No hyphenation pattern has been}%
\typeout{** loaded for the language `#1'. Using the pattern for}%
\typeout{** the default language instead.}%
\else
\language=\csname l@#1\endcsname
\fi
#2}}
\providecommand{\BIBdecl}{\relax}
\BIBdecl

\bibitem{Zhang1}
L.~Bai, W.~Zhang, X.~Pan, and C.~Zhao, ``Underwater image enhancement based on global and local equalization of histogram and dual-image multi-scale fusion,'' \emph{IEEE Access}, vol.~8, pp. 128\,973--128\,990, 2020.

\bibitem{ahmed2023vision}
M.~Ahmed, A.~B. Bakht, T.~Hassan, W.~Akram, A.~Humais, L.~Seneviratne, S.~He, D.~Lin, and I.~Hussain, ``Vision-based autonomous navigation for unmanned surface vessel in extreme marine conditions,'' in \emph{2023 IEEE/RSJ International Conference on Intelligent Robots and Systems (IROS)}.\hskip 1em plus 0.5em minus 0.4em\relax IEEE, 2023, pp. 7097--7103.

\bibitem{guo2019underwater}
Y.~Guo, H.~Li, and P.~Zhuang, ``Underwater image enhancement using a multiscale dense generative adversarial network,'' \emph{IEEE Journal of Oceanic Engineering}, vol.~45, no.~3, pp. 862--870, 2019.

\bibitem{azmi2019natural}
K.~Z.~M. Azmi, A.~S.~A. Ghani, Z.~M. Yusof, and Z.~Ibrahim, ``Natural-based underwater image color enhancement through fusion of swarm-intelligence algorithm,'' \emph{Applied Soft Computing}, vol.~85, p. 105810, 2019.

\bibitem{raveendran2021underwater}
S.~Raveendran, M.~D. Patil, and G.~K. Birajdar, ``Underwater image enhancement: a comprehensive review, recent trends, challenges and applications,'' \emph{Artificial Intelligence Review}, vol.~54, pp. 5413--5467, 2021.

\bibitem{anwar2020diving}
S.~Anwar and C.~Li, ``Diving deeper into underwater image enhancement: A survey,'' \emph{Signal Processing: Image Communication}, vol.~89, p. 115978, 2020.

\bibitem{d2}
C.~Fabbri, M.~J. Islam, and J.~Sattar, ``Enhancing underwater imagery using generative adversarial networks,'' in \emph{2018 IEEE International Conference on Robotics and Automation (ICRA)}, 2018, pp. 7159--7165.

\bibitem{d1}
J.~Li, K.~A. Skinner, R.~M. Eustice, and M.~Johnson-Roberson, ``Watergan: Unsupervised generative network to enable real-time color correction of monocular underwater images,'' \emph{IEEE Robotics and Automation Letters}, vol.~3, no.~1, pp. 387--394, 2018.

\bibitem{d5}
M.~J. Islam, Y.~Xia, and J.~Sattar, ``Fast underwater image enhancement for improved visual perception,'' \emph{IEEE Robotics and Automation Letters}, vol.~5, no.~2, pp. 3227--3234, 2020.

\bibitem{d6}
C.~Li, S.~Anwar, and F.~Porikli, ``Underwater scene prior inspired deep underwater image and video enhancement,'' \emph{Pattern Recognition}, vol.~98, p. 107038, 2020.

\bibitem{d7}
C.~Li, C.~Guo, W.~Ren, R.~Cong, J.~Hou, S.~Kwong, and D.~Tao, ``An underwater image enhancement benchmark dataset and beyond,'' \emph{IEEE Transactions on Image Processing}, vol.~29, pp. 4376--4389, 2019.

\bibitem{hummel1975image}
R.~Hummel, ``Image enhancement by histogram transformation,'' \emph{Unknown}, 1975.

\bibitem{reza2004realization}
A.~M. Reza, ``Realization of the contrast limited adaptive histogram equalization (clahe) for real-time image enhancement,'' \emph{Journal of VLSI signal processing systems for signal, image and video technology}, vol.~38, pp. 35--44, 2004.

\bibitem{karam2013enhancement8}
G.~S. Karam, Z.~M. Abood, and R.~N. Saleh, ``Enhancement of underwater image using fuzzy histogram equalization,'' \emph{International Journal of Applied Information Systems}, vol.~6, no.~6, pp. 1--6, 2013.

\bibitem{petit2009underwater9}
F.~Petit, A.-S. Capelle-Laiz{\'e}, and P.~Carr{\'e}, ``Underwater image enhancement by attenuation inversionwith quaternions,'' in \emph{2009 IEEE International Conference on Acoustics, Speech and Signal Processing}.\hskip 1em plus 0.5em minus 0.4em\relax IEEE, 2009, pp. 1177--1180.

\bibitem{ancuti2012enhancing10}
C.~Ancuti, C.~O. Ancuti, T.~Haber, and P.~Bekaert, ``Enhancing underwater images and videos by fusion,'' in \emph{2012 IEEE conference on computer vision and pattern recognition}.\hskip 1em plus 0.5em minus 0.4em\relax IEEE, 2012, pp. 81--88.

\bibitem{he2010single11}
K.~He, J.~Sun, and X.~Tang, ``Single image haze removal using dark channel prior,'' \emph{IEEE transactions on pattern analysis and machine intelligence}, vol.~33, no.~12, pp. 2341--2353, 2010.

\bibitem{drews2016underwater12}
P.~L. Drews, E.~R. Nascimento, S.~S. Botelho, and M.~F.~M. Campos, ``Underwater depth estimation and image restoration based on single images,'' \emph{IEEE computer graphics and applications}, vol.~36, no.~2, pp. 24--35, 2016.

\bibitem{peng2018generalization13}
Y.-T. Peng, K.~Cao, and P.~C. Cosman, ``Generalization of the dark channel prior for single image restoration,'' \emph{IEEE Transactions on Image Processing}, vol.~27, no.~6, pp. 2856--2868, 2018.

\bibitem{chang2018single14}
H.-H. Chang, C.-Y. Cheng, and C.-C. Sung, ``Single underwater image restoration based on depth estimation and transmission compensation,'' \emph{IEEE Journal of Oceanic Engineering}, vol.~44, no.~4, pp. 1130--1149, 2018.

\bibitem{perez2017benchmarking15}
J.~Perez, P.~J. Sanz, M.~Bryson, and S.~B. Williams, ``A benchmarking study on single image dehazing techniques for underwater autonomous vehicles,'' in \emph{OCEANS 2017-Aberdeen}.\hskip 1em plus 0.5em minus 0.4em\relax IEEE, 2017, pp. 1--9.

\bibitem{mcglamery1975computer16}
B.~McGlamery, ``Computer analysis and simulation of underwater camera system performance,'' \emph{SIO ref}, vol.~75, no.~2, 1975.

\bibitem{mcglamery1980computer17}
------, ``A computer model for underwater camera systems,'' in \emph{Ocean Optics VI}, vol. 208.\hskip 1em plus 0.5em minus 0.4em\relax SPIE, 1980, pp. 221--231.

\bibitem{akkaynak2017space18}
D.~Akkaynak, T.~Treibitz, T.~Shlesinger, Y.~Loya, R.~Tamir, and D.~Iluz, ``What is the space of attenuation coefficients in underwater computer vision?'' in \emph{Proceedings of the IEEE Conference on Computer Vision and Pattern Recognition}, 2017, pp. 4931--4940.

\bibitem{akkaynak2018revised19}
D.~Akkaynak and T.~Treibitz, ``A revised underwater image formation model,'' in \emph{Proceedings of the IEEE conference on computer vision and pattern recognition}, 2018, pp. 6723--6732.

\bibitem{akkaynak2019sea20}
------, ``Sea-thru: A method for removing water from underwater images,'' in \emph{Proceedings of the IEEE/CVF conference on computer vision and pattern recognition}, 2019, pp. 1682--1691.

\bibitem{d3}
J.-Y. Zhu, T.~Park, P.~Isola, and A.~A. Efros, ``Unpaired image-to-image translation using cycle-consistent adversarial networks,'' in \emph{Proceedings of the IEEE international conference on computer vision}, 2017, pp. 2223--2232.

\bibitem{d4}
Y.~Guo, H.~Li, and P.~Zhuang, ``Underwater image enhancement using a multiscale dense generative adversarial network,'' \emph{IEEE Journal of Oceanic Engineering}, vol.~45, no.~3, pp. 862--870, 2019.

\bibitem{d8}
M.~J. Islam, P.~Luo, and J.~Sattar, ``Simultaneous enhancement and super-resolution of underwater imagery for improved visual perception,'' \emph{arXiv preprint arXiv:2002.01155}, 2020.

\bibitem{d9}
Q.~Qi, K.~Li, H.~Zheng, X.~Gao, G.~Hou, and K.~Sun, ``Sguie-net: Semantic attention guided underwater image enhancement with multi-scale perception,'' \emph{IEEE Transactions on Image Processing}, vol.~31, pp. 6816--6830, 2022.

\bibitem{d10}
Z.~Huang, J.~Li, Z.~Hua, and L.~Fan, ``Underwater image enhancement via adaptive group attention-based multiscale cascade transformer,'' \emph{IEEE Transactions on Instrumentation and Measurement}, vol.~71, pp. 1--18, 2022.

\bibitem{d11}
X.~Cai, N.~Jiang, W.~Chen, J.~Hu, and T.~Zhao, ``Cure-net: A cascaded deep network for underwater image enhancement,'' \emph{IEEE Journal of Oceanic Engineering}, 2023.

\bibitem{zhang2018image}
Y.~Zhang, K.~Li, K.~Li, L.~Wang, B.~Zhong, and Y.~Fu, ``Image super-resolution using very deep residual channel attention networks,'' in \emph{Proceedings of the European conference on computer vision (ECCV)}, 2018, pp. 286--301.

\bibitem{zhang2019residual}
Y.~Zhang, K.~Li, K.~Li, B.~Zhong, and Y.~Fu, ``Residual non-local attention networks for image restoration,'' \emph{arXiv preprint arXiv:1903.10082}, 2019.

\bibitem{zamir2022}
S.~W. Zamir, A.~Arora, S.~H. Khan, H.~Munawar, F.~S. Khan, M.-H. Yang, and L.~Shao, ``Learning enriched features for fast image restoration and enhancement,'' \emph{IEEE Transactions on Pattern Analysis and Machine Intelligence}, 2022.

\bibitem{s&e}
J.~Hu, L.~Shen, and G.~Sun, ``Squeeze-and-excitation networks,'' in \emph{2018 IEEE/CVF Conference on Computer Vision and Pattern Recognition}, 2018, pp. 7132--7141.

\bibitem{CBAM}
S.~Woo, J.~Park, J.-Y. Lee, and I.~S. Kweon, ``Cbam: Convolutional block attention module,'' in \emph{Proceedings of the European conference on computer vision (ECCV)}, 2018, pp. 3--19.

\bibitem{PatchGAN}
P.~Isola, J.-Y. Zhu, T.~Zhou, and A.~A. Efros, ``Image-to-image translation with conditional adversarial networks,'' in \emph{2017 IEEE Conference on Computer Vision and Pattern Recognition (CVPR)}, 2017, pp. 5967--5976.

\bibitem{yi2017dualgan}
Z.~Yi, H.~Zhang, P.~Tan, and M.~Gong, ``Dualgan: Unsupervised dual learning for image-to-image translation,'' in \emph{Proceedings of the IEEE international conference on computer vision}, 2017, pp. 2849--2857.

\bibitem{Perceptual}
J.~Johnson, A.~Alahi, and L.~Fei-Fei, ``Perceptual losses for real-time style transfer and super-resolution,'' in \emph{Computer Vision--ECCV 2016: 14th European Conference, Amsterdam, The Netherlands, October 11-14, 2016, Proceedings, Part II 14}.\hskip 1em plus 0.5em minus 0.4em\relax Springer, 2016, pp. 694--711.

\bibitem{VGG19}
K.~Simonyan and A.~Zisserman, ``Very deep convolutional networks for large-scale image recognition,'' \emph{arXiv preprint arXiv:1409.1556}, 2014.

\bibitem{li2019fusion}
H.~Li, J.~Li, and W.~Wang, ``A fusion adversarial underwater image enhancement network with a public test dataset,'' \emph{arXiv preprint arXiv:1906.06819}, 2019.

\bibitem{8949763}
R.~Liu, X.~Fan, M.~Zhu, M.~Hou, and Z.~Luo, ``Real-world underwater enhancement: Challenges, benchmarks, and solutions under natural light,'' \emph{IEEE Transactions on Circuits and Systems for Video Technology}, vol.~30, no.~12, pp. 4861--4875, 2020.

\bibitem{PSNR}
A.~Horé and D.~Ziou, ``Image quality metrics: Psnr vs. ssim,'' in \emph{2010 20th International Conference on Pattern Recognition}, 2010, pp. 2366--2369.

\bibitem{SSIM}
Z.~Wang, A.~C. Bovik, H.~R. Sheikh, and E.~P. Simoncelli, ``Image quality assessment: from error visibility to structural similarity,'' \emph{IEEE transactions on image processing}, vol.~13, no.~4, pp. 600--612, 2004.

\bibitem{UIQM}
K.~Panetta, C.~Gao, and S.~Agaian, ``Human-visual-system-inspired underwater image quality measures,'' \emph{IEEE Journal of Oceanic Engineering}, vol.~41, no.~3, pp. 541--551, 2016.

\bibitem{UCIQE}
M.~Yang and A.~Sowmya, ``An underwater color image quality evaluation metric,'' \emph{IEEE Transactions on Image Processing}, vol.~24, no.~12, pp. 6062--6071, 2015.

\bibitem{NIQE}
A.~Mittal, R.~Soundararajan, and A.~C. Bovik, ``Making a “completely blind” image quality analyzer,'' \emph{IEEE Signal Processing Letters}, vol.~20, no.~3, pp. 209--212, 2013.

\bibitem{Fusion}
C.~O. Ancuti, C.~Ancuti, C.~De~Vleeschouwer, and P.~Bekaert, ``Color balance and fusion for underwater image enhancement,'' \emph{IEEE Transactions on Image Processing}, vol.~27, no.~1, pp. 379--393, 2018.

\bibitem{IBLA}
Y.-T. Peng and P.~C. Cosman, ``Underwater image restoration based on image blurriness and light absorption,'' \emph{IEEE transactions on image processing}, vol.~26, no.~4, pp. 1579--1594, 2017.

\bibitem{U-Trans}
L.~Peng, C.~Zhu, and L.~Bian, ``U-shape transformer for underwater image enhancement,'' \emph{IEEE Transactions on Image Processing}, 2023.

\bibitem{8099551}
D.~Akkaynak, T.~Treibitz, T.~Shlesinger, Y.~Loya, R.~Tamir, and D.~Iluz, ``What is the space of attenuation coefficients in underwater computer vision?'' in \emph{2017 IEEE Conference on Computer Vision and Pattern Recognition (CVPR)}, 2017, pp. 568--577.

\bibitem{akram2024aquaculture}
W.~Akram, T.~Hassan, H.~Toubar, M.~Ahmed, N.~Mi{\v{s}}kovic, L.~Seneviratne, and I.~Hussain, ``Aquaculture defects recognition via multi-scale semantic segmentation,'' \emph{Expert Systems with Applications}, vol. 237, p. 121197, 2024.

\bibitem{akram2022visual}
W.~Akram, A.~Casavola, N.~Kapetanovi{\'c}, and N.~Mi{\v{s}}kovic, ``A visual servoing scheme for autonomous aquaculture net pens inspection using rov,'' \emph{Sensors}, vol.~22, no.~9, p. 3525, 2022.

\end{thebibliography}

\smallskip

\end{document}